\documentclass[10pt, a4paper]{article}

\usepackage[final]{LREC2026/lrec2026} 
\usepackage{multirow}
\usepackage{booktabs}
\usepackage{makecell} 

\title{Introducing A Bangla Sentence – Gloss Pair Dataset for Bangla Sign Language Translation and Research}

\name{
  \begin{tabular}{c}
  Neelavro Saha\textsuperscript{1}, Rafi Shahriyar\textsuperscript{1}, Nafis Ashraf Roudra\textsuperscript{1},\\
  Saadman Sakib\textsuperscript{1}, Annajiat Alim Rasel\textsuperscript{1}
  \end{tabular}
}

\address{
\textsuperscript{1}\textit{Department of Computer Science and Engineering, BRAC University, Bangladesh}\\
\begin{tabular}{c}
neelavro.saha@g.bracu.ac.bd, rafi.shahriyar@g.bracu.ac.bd, nafis.ashraf@g.bracu.ac.bd,\\
saadman.sakib1@g.bracu.ac.bd, annajiat@bracu.ac.bd
\end{tabular}\\
}

\abstract{
Bangla Sign Language (BdSL) translation represents a low-resource NLP task due to the lack of large-scale datasets that address sentence-level translation. Correspondingly, existing research in this field has been limited to word and alphabet level detection. In this work, we introduce Bangla-SGP, a novel parallel dataset consisting of 1,000 human-annotated sentence–gloss pairs which was augmented with around 3,000 synthetically generated pairs using syntactic and morphological rules through a rule-based Retrieval-Augmented Generation (RAG) pipeline. The gloss sequences of the spoken Bangla sentences are made up of individual glosses which are Bangla sign supported words and serve as an intermediate representation for a continuous sign. Our dataset consists of 1000 high quality Bangla sentences that are manually annotated into a gloss sequence by a professional signer. The augmentation process incorporates rule-based linguistic strategies and prompt engineering techniques that we have adopted by critically analyzing our human annotated sentence-gloss pairs and by working closely with our professional signer. Furthermore, we fine-tune several transformer-based models such as mBart50, Google mT5, GPT4.1-nano and evaluate their sentence-to-gloss translation performance using BLEU scores, based on these evaluation metrics we compare the model's gloss-translation consistency across our dataset and the RWTH-PHOENIX-2014T benchmark.
 \\ \newline  }

\begin{document}

\maketitleabstract

\section{Introduction}
Developments in Natural Language Processing has made multiple spoken languages more accessible through machine translation. Following the introduction of Transformer based architectures, models can be trained to exhibit enhanced multilingual understanding, allowing information to be distributed despite linguistic barriers \cite{roberta,liu-etal-2020-multilingual-denoising}. However, even in today’s world, there still remains a huge accessibility challenge for the deaf community - particularly when it comes to reading and understanding text or spoken language. Deaf people prefer sign language as a primary communication method. However, this mode of communication is not incorporated in much of the content of the modern day, which are mostly voice or text-based. Research shows that when people read, they associate words with their corresponding pronunciations in order to comprehend the text \citep{reading-and-deafness}. Deaf people, unable to hear, lack this phonological skill as they have no idea what the words sound like. As such, reading text is challenging for the deaf population. To tackle these issues, ongoing research include creating pipelines of different deep-learning and transformer based models to convert spoken text to its corresponding sign language translation in an animated form \cite{sentencetogloss1, zuo2024simplebaselinespokenlanguage}. These pipelines require extensive datasets containing spoken sentences, their corresponding videos, and gloss sequences that serve as a type of structured annotations that represent sign language in textual form. While text-to-gloss translation has been explored in various languages globally, little work has been done in the context of Bangla. In Bangladesh, there are approximately 13 million people with hearing difficulties, 3 million of whom are deaf \citep{Alauddin2004}. Many of these people use Bangla sign language (BdSL) as the primary method of communication. Therefore, in this paper we present the following contributions: Firstly, we introduce Bangla-SGP, the first phase of our BdSL dataset, which could serve as a foundational step for facilitating research in Continuous Bangla Sign Language Recognition and translation, a field that remains vastly unexplored due to a severe lack of publicly available resources. Secondly, we propose a novel rule-based RAG pipeline for Bangla sentence-gloss pair generation, which helps address the low-resource challenge in BdSL by partially automating the gloss generation process and reducing reliance on extensive manual annotation. Finally, our generated gloss sequences can also be used as an intermediary step toward 3D animated representation of sign language sentences, enabling real-world applications such as education tools and virtual interpreters to support the deaf community.
    
    


\section{Related Works}

\subsection{Bangladeshi Sign Language (BdSL)}

BdSL60, an extensive and one of the very few word-level BdSL dataset consisting of 60 annotated Bangla sign words and 9307 video trials from 18 signers \citep{iut}. Various preprocessing methods were applied to the dataset. A PROLONGED version of the dataset was created by duplicating frames to maintain uniformity for machine learning models. A FLIPPED version was also created by transforming all left-handed sign instances to right-handed. A Relative Quantization (RQ) dataset was also created by quantizing the landmark points to reduce depth variations. Overall, the dataset variations boil down to: i) PROLONGED and FLIPPED, ii) PROLONGED and NON-FLIPPED, iii) NON-PROLONGED and FLIPPED, iii) NON-PROLONGED and NON-FLIPPED and finally v) RQ. Mediapipe Hollistic was used to extract pose keypoints and traditional SVM, SVM-DTW, and Attention-based Bi-LSTM were used to benchmark the dataset. Classical SVM provided a testing accuracy of about 67.5\%, while SVM-DTW scored around 65.8\%. On the other hand, attention-based biLSTM had a testing accuracy of up to 75.1\%.  

\citet{10306914} introduced SignBD-Word a word-level dataset in video format for Bangla sign language. The dataset contains 6,000 video clips of 200 distinct Bangla sign language words with each sign recorded from both full-body and upper-body perspectives. Additionally, annotations and glosses are provided for each video. Collected with the help of 16 different signers, SignBD-word is a dataset that serves as a baseline for future research in BdSL recognition and pose synthesis. To standardize the dimensions of all sign language video clips, individual videos were sampled at 30 frames and  at consistent intervals. Individual frames were then re-scaled to 224×224 pixels using bi-linear interpolation. They also compare 2D CNN with LSTM and 3D CNN approaches to determine which architectures best capture the spatio-temporal dynamics of sign gestures. 3D CNN models such as I3D and SlowFast outperform 2D CNN models significantly. Subsequently the authors explored GAN-based methods for synthesizing sign languages videos from the 2D body-pose skeletons. Three models were used they are as follows, CycleGan, pix2pixHD and SPADE. CycleGAN model produced blurred and noisy results correspondingly SPADE also struggled due to its reliance on semantic label maps. Therefore, pix2pixHD stood out as the best as outputs consisted of clear hand and face details. Realism of the synthesized videos were determined through a Mean Opinion Score (MOS) test. Here  20 university students rated the generated videos. Out of the 3 models pix2pixHD was rated the highest. The primary limitations of this study includes the limited number of signers, the restricted set of sign words, and the need for further fine-tuning of the GAN models for better performance.

\subsection{Sign Language Datasets}

\citet{ms-asl} introduced a large dataset for American Sign Language (ASL) recognition for training robust deep learning models . Sign language recognition is a challenging task due to its multimodal nature which involves continuous hand movement, different body orientations, handshapes and facial expressions. As such, a large scale dataset is required to enable deep learning techniques for sign language recognition. This was the motivation behind this paper as it tries to bridge the gap between ASL recognition and current computer vision advancements. The dataset, MS-ASL, has over 25 thousand videos of approximately 1000 ASL signs. Moreover, it has 222 distinct signers, and the videos were shot in an uncontrolled environment, under real-life conditions with variations in lighting and background. These videos were collected from public sources like educational ASL videos from YouTube. Subsequently, the signs were processed to segment the signs into individual samples. OCR was used to extract text from the videos and the subtitles were used for metadata. Then it was reviewed manually to ensure quality, particularly by cropping the videos that are too long. Also, synonyms with ASL vocabulary were handled manually. Afterwards the dataset was divided into train, test and validation sets containing videos of 165, 20 and 37 signers respectively, and state of the art models like I3D, 2D-CNN-LSTM and body key-point recognition were used to benchmark the dataset. Among these models, I3D had the best results as it showed an accuracy of 57.69\%, while the other models had much lower accuracy. I3D also had the best results in the top-five accuracy results, as it scored 81.08\% in that metric. While this large scale dataset offers significant advancements in sign language recognition, it has some limitations. There were not sufficient samples for some of the signs which resulted in these classes having lower performance. Additionally, some non-native annotators were involved in the manual annotation process which may introduce some errors. Furthermore, the dataset does not address regional variations in ASL, which may affect the generalizability of the models.

\section{Methodology}

\subsection{Dataset Creation}
\begin{figure}[h!]
  \centering
  \includegraphics[width=\linewidth]{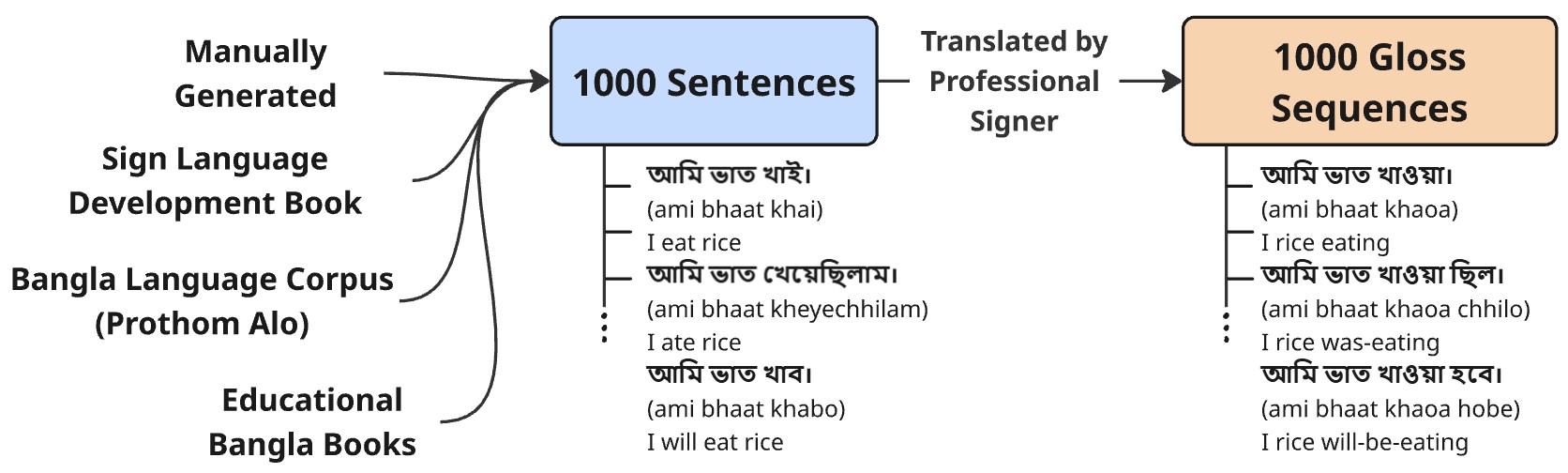}
  \caption{Dataset Creation}
  \label{dataset_creation}
\end{figure}

Firstly, we created a foundation dataset of 1000 Bangla sentence-gloss pairs. As shown in Figure~\ref{dataset_creation}, these sentences were collected from a variety of sources, including Bangla newspaper corpora, bangla literature, BdSL language development book etc. The sentences were annotated to their corresponding gloss by Mr. Ariful Islam who is an experienced Bangla sign language presenter at BTV.

\subsection{Data Augmentation}

\begin{figure}[h!]
  \centering
  \includegraphics[width=\linewidth]{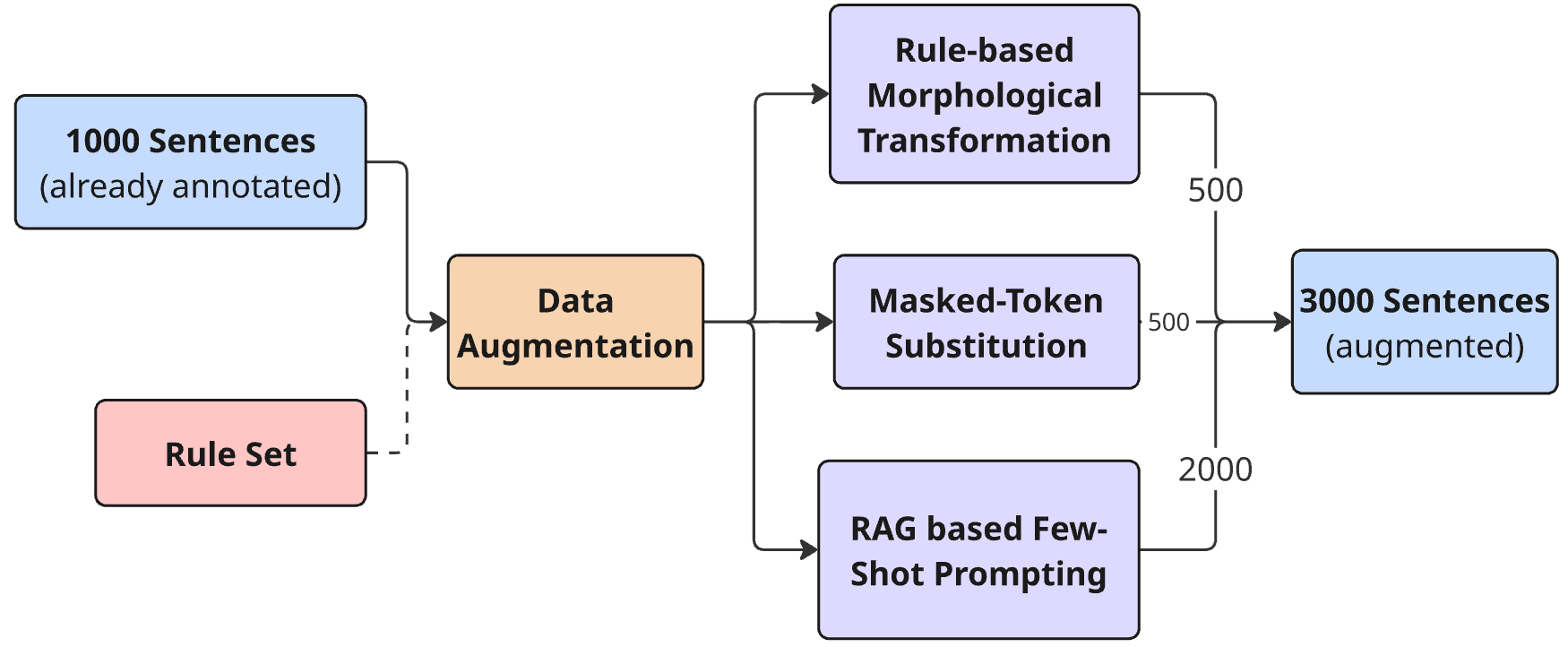}
  \caption{Data Augmentation}
  \label{fig:rag_augmentation}
\end{figure}

Data augmentation plays a crucial role in addressing data scarcity, especially in low-resource language tasks where annotated data is limited. Figure~\ref{fig:rag_augmentation} illustrates the workflow of our data augmentation process. Specifically in our case of Bangla Text-to-BdSL gloss translation, we curated a dataset of 1000 sentence-gloss pairs annotated by a professional signer. While this dataset is larger than other similar works in the field of Bangla text to gloss translation, it still remains insufficient in achieving satisfactory results in model training. As such, we applied several data augmentation strategies to enhance the dataset which include: (1) Rule-based Morphological Transformation, (2) Masked-Token Substitution, (3) Retrieval-Augmented Generation (RAG). With these methods, we generated a total of 3000 new entries of text-gloss translations. Our final dataset now has 4000 entries with 1:3 ratio for original to augmentation.

\subsubsection{Rule-based Morphological Transformation}

\begin{figure}[h!]
  \centering
  \includegraphics[width=\linewidth]{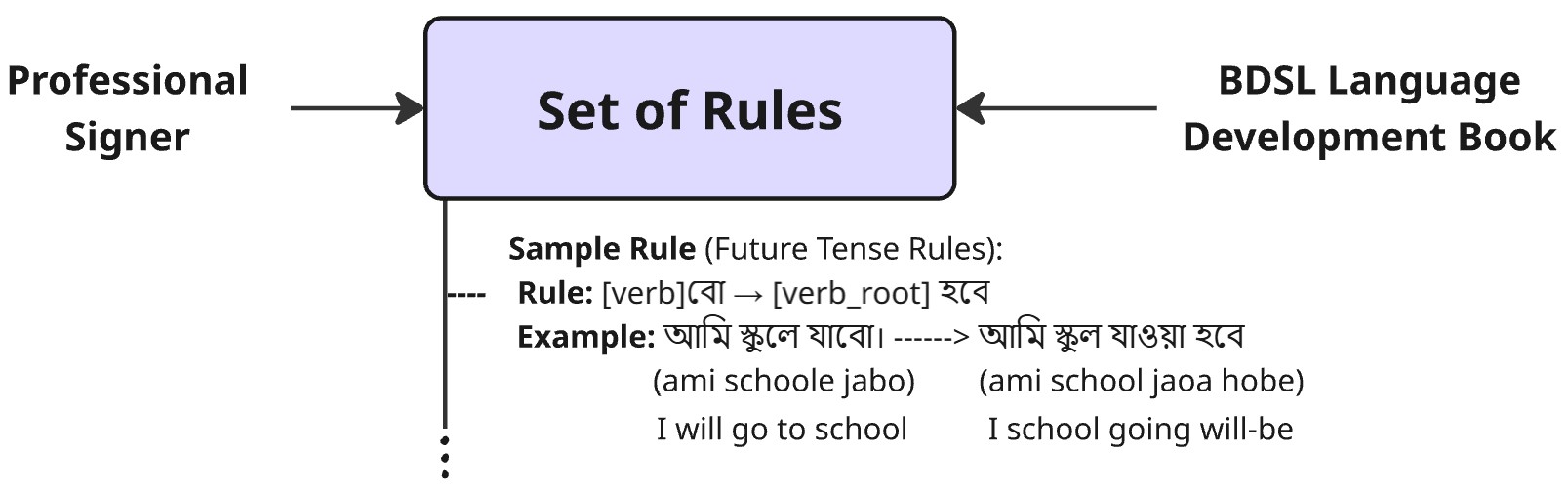}
  \caption{Rule Set Collection}
  \label{rule_set_collection}
\end{figure}

We collaborated closely with a professional signer to analyze and extract common translation patterns from our annotated dataset (Figure~\ref{rule_set_collection}). A significant portion of these patterns were related to tense. Each tense follows consistent morphological rules that map the verb of the text form into gloss. For example, a common pattern in the future tense is that the verb is translated to its root form followed by the equivalent of “will be.” Based on such consistent mappings, we, along with the help of the professional signer, compiled a set of rules. Afterwards, we applied these rules to augment existing sentences by systematically converting them into other tenses.

\begin{figure}[h!]
  \centering
  \includegraphics[width=\linewidth]{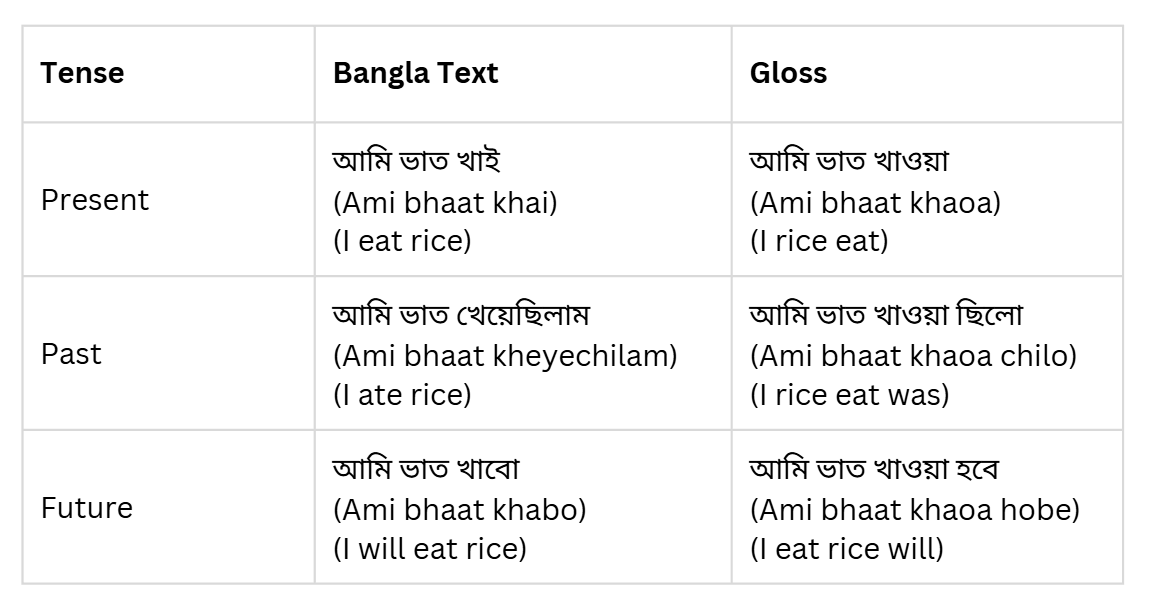}
  \caption{Example of verb tense transformations used in rule-based morphological augmentation.}
  \label{fig:morphological_examples}
\end{figure}

An example of this transformation process is illustrated in Figure~\ref{fig:morphological_examples}. This method ensured that the augmented sentence-gloss pairs remained linguistically accurate and semantically faithful to the original meaning. Furthermore, because the glosses were generated using verified rules rather than arbitrary transformations, the resulting data maintained a high level of consistency. With this, we manually generated around 500 text-gloss pairs and expanded the dataset in a controlled and interpretable manner, which is particularly important in low-resource tasks such as Bangla text-to-gloss translation.

\subsubsection{Masked-Token Substitution}

\begin{figure*}[t!]
  \centering
  \includegraphics[width=\textwidth]{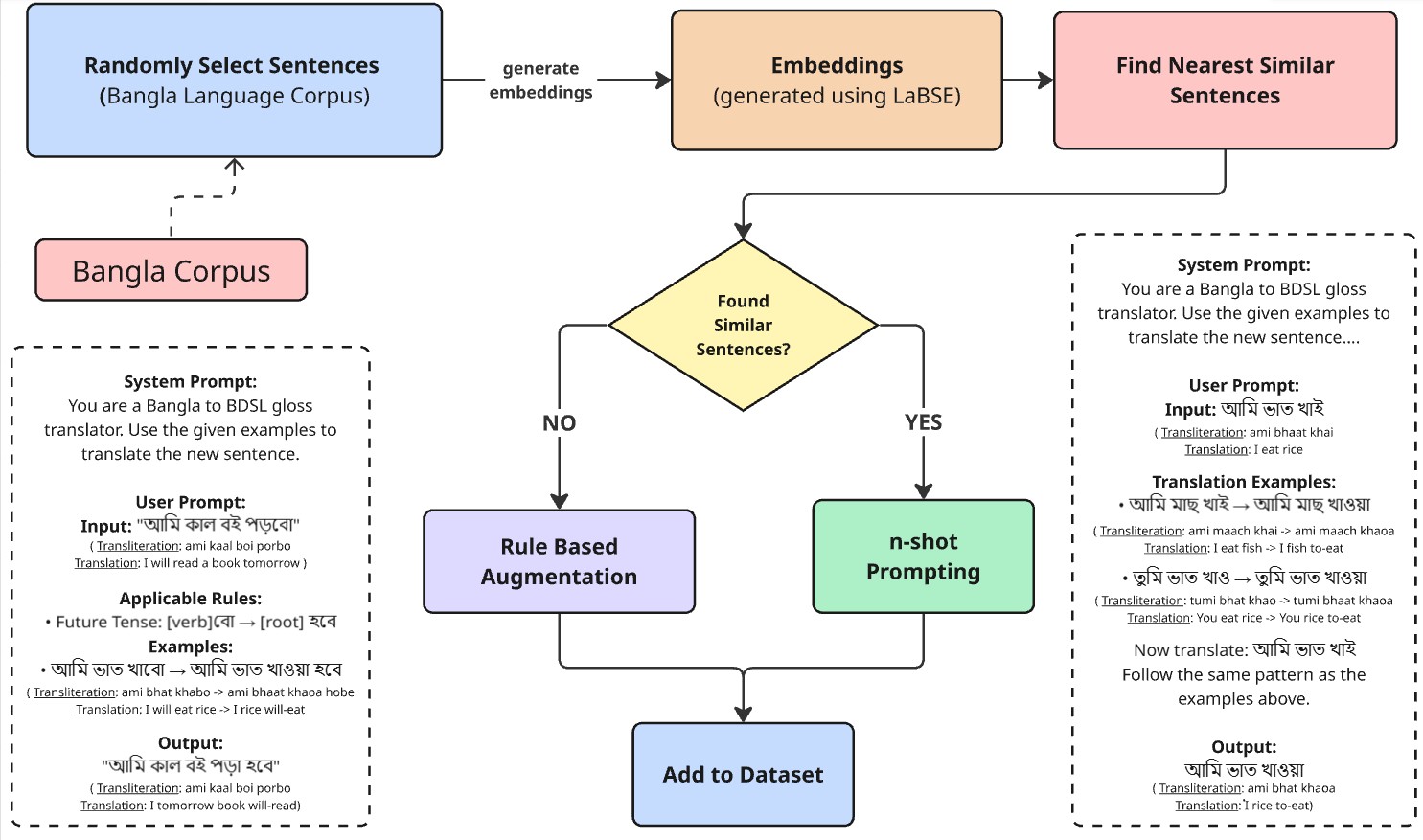}
  \caption{RAG Based Augmentation}
  \label{RAG Based Augmentation}
\end{figure*}

The second method we applied is a form of contextual augmentation known as Masked Token Substitution. This technique creates lexical variants of sentence-gloss pairs using known rules and the Bangla-BERT model. We defined templates where certain words (typically common nouns or verbs) were masked. Bangla-BERT then predicted the most contextually appropriate replacements for the masked tokens. The newly generated sentences preserved the grammatical structure and meaning of the originals.

\begin{figure}[h!]
  \centering
  \includegraphics[width=\linewidth]{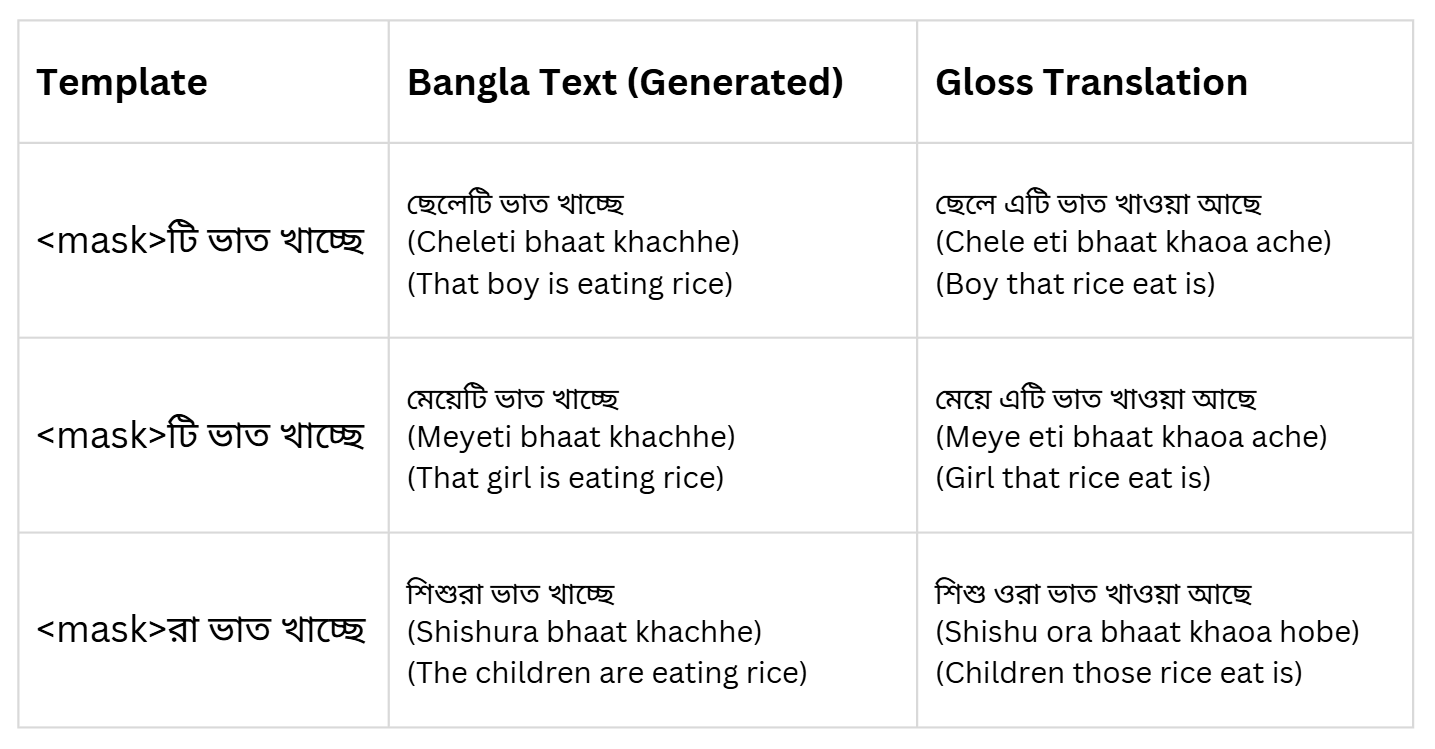}
  \caption{Example of masked token substitution where Bangla-BERT generates contextually valid replacements for masked tokens, producing diverse yet grammatically consistent sentence variants.}
  \label{fig:masking_examples}
\end{figure}

An illustrative example of this approach is shown in Figure~\ref{fig:masking_examples}. By generating diverse sentence forms through this method, we aimed to expose the model to a broader range of linguistic patterns, improving its generalization capability while maintaining the rules of text-to-gloss translation. Approximately 500 text-gloss pairs were created using this approach.

\subsubsection{Retrieval-Augmented Generation (RAG) based Few-Shot Prompting}


Our third approach utilized a Retrieval-Augmented Generation (RAG) pipeline for data augmentation. We split the 1000 professionally annotated text-gloss samples into 80\%, 10\%, and 10\% for training, testing, and validation respectively. The training set was embedded using LaBSE and stored in Pinecone, a vector database. This training data served as the ground truth reference for generating additional synthetic samples. For retrieving supplementary Bangla text samples, we used the publicly available Bangla Corpus by Nuhash Afnan, which contains sentences collected from newspapers on varied topics. We selected this corpus because the formal writing style of newspaper closely aligns with the professionally annotated training data used in our experiments. Figure \ref{RAG Based Augmentation} illustrates the workflow of our RAG based augmentation process. We used GPT-4.1-nano for few-shot prompting and a custom two-stage-prompting for gloss generation. 
 
 Retrieval-Augmented Generation (RAG), paired with few-shot prompting, is particularly effective for data augmentation in low-resource settings because it grounds a model’s generation by providing contextually similar examples to learn from. Particularly in our case of Bangla text-to-gloss translation, it enables the model to produce more accurate translations even for some rare or less frequent sentence structures that are not covered by the common rule sets we compiled. For example, some words do not have direct translations in BDSL. For instance, the word ``recipe'' does not have a corresponding sign and is instead translated to 
(khabar ranna niyom) 'food cooking rules'. Without giving such context during data augmentation using GPT, these patterns may not show up in the output. As such, we implemented a RAG based approach that first retrieves the similar occurrences from our professionally annotated dataset. This way, if an input sentence contains a word like ``recipe'', 
the system is likely to retrieve examples containing similar or same terms. These retrieved examples, along with their corresponding glosses, are then included in a prompt passed as a few shot prompt. The model, having access to these grounded examples, is better guided to produce an accurate and contextually appropriate gloss for the new sentence. 

We set a similarity score threshold of 0.5, meaning that if the score is above 0.5, that is considered a similar sentence usable for few shot prompt.  To prevent overload, we also cap the number of retrieved examples at 20 per prompt. Now the problem arises if less or no similar sentences are found. To address this, we implement a fallback mechanism where, instead of sending n similar sentences, it sends the rules in the prompt. We further improved this approach by implementing a two-stage prompting strategy. Instead of sending all the rules at once, which may overwhelm the model due to the complexity and volume of the input, we first issue a prompt to identify the tense of the sentence. Then, based on the tense, we only send the rules of that particular tense along with the examples. This way the prompt is more focused and likely to generate better results. This two-stage prompting strategy draws on principles similar to chain-of-thought prompting, where an intermediate reasoning step (tense identification in our case) is used to refine the final generation prompt. We generated 2000 text-gloss pairs using this approach, and applied the Cohen’s kappa method to validate our results.

 \subsection{Validation using Cohen's kappa}
 
 Cohen’s kappa is a widely used statistical measure for evaluating how consistently different annotators label the same data. It is particularly useful in validating data augmentation techniques in low-resource settings where ground truth data is limited. In our case, we first generated 1000 text-gloss pairs using our RAG based approach. Then, we randomly selected 15\% (150 samples) of the text-gloss pairs and asked two professional BDSL signers to independently review and validate the gloss translations. The signers who validated these translations are Mr. Ariful Islam and Mrs. Tanjila Tartushi, both of whom are professional sign presenters at BTV. They reviewed the samples independently and did not consult each other. They were given two input fields to validate each entry. The first field was a Yes/ No question asking if they think the gloss is understandable. The second field asked them to rate the gloss on a scale of 1–5, where 1 indicated the least understandable and most inaccurate translation, and 5 indicated most accurate translation. We then calculated the Cohen’s kappa score between the two signers’ judgments.

\begin{table}[h!]
\centering
\resizebox{\columnwidth}{!}{
\begin{tabular}{lccc}
\toprule
\textbf{Metric} & \textbf{Signer 1} & \textbf{Signer 2} & \textbf{Combined} \\
\midrule
Validation Rate (\%) & 74.7 & 76.0 & 75.3 \\
Average Quality Score & 2.96 & 3.41 & 3.19 \\
\midrule
\multicolumn{4}{l}{\textbf{Quality Distribution}} \\
High Quality (Score $\geq$ 4) & 35.3\% & 50.0\% & 42.7\% \\
Acceptable (Score = 3) & 26.0\% & 22.7\% & 24.3\% \\
Low Quality (Score $\leq$ 2) & 38.7\% & 27.3\% & 33.0\% \\
\midrule
\multicolumn{4}{l}{\textbf{Inter-rater Reliability (Cohen's Kappa)}} \\
Binary Agreement & \multicolumn{3}{c}{$\kappa = 0.7489$ (Substantial)} \\
Quality Agreement & \multicolumn{3}{c}{$\kappa = 0.3496$ (Fair)} \\
\bottomrule
\end{tabular}
}
\caption{Key evaluation metrics for RAG-based data augmentation.}
\label{tab:key_metrics}
\end{table}

Table \ref{tab:key_metrics} shows the key evaluation metrics for the effectiveness of our RAG-based data augmentation method. It shows that the validation rate, which measures the percentage of glosses marked as correct by professional signers, was 74.7\% for Signer 1 and 76.0\% for Signer 2 and the average validation rate is 75.3\%. The average quality score, based on a 1–5 scale, was 2.96 from Signer 1 and 3.41 from Signer 2 and an average of 3.19 out of 5. The quality distribution section breaks down how many glosses were rated as high quality, acceptable, or low quality (score $\leq$ 2). Overall, 42.7\% of glosses were rated as high quality and 24.3\% were acceptable, indicating that most glosses met practical quality standards. Lastly, the reliability is measured using Cohen’s kappa, which shows substantial agreement ($\kappa = 0.7489$) for binary validation and fair agreement  ($\kappa = 0.3496$)  for quality ratings. This supports our conclusion that the evaluations were consistent and trustworthy. As such, we can confidently use the augmented data as a reliable resource for Bangla text-to-gloss translation tasks.

\section{Experiments}

This following work investigates the leveraging of four separate transformer-based models, mBART-50, NLLB-200, mT5 as well as GPT-4.1-nano for translating Bengali-to-Bangla Sign Language glosses, centered around a low-resource task. These models all consist of an encoder-decoder architecture and are pretrained with multiple language corpora, ensuring their suitability in low-resource data cases. The original dataset of 1,000 expert-annotated sentence-gloss pairs was split into training, validation, and test sets in the proportions of 80\%, 10\%, and 10\%, respectively. The same test set was used when evaluating the augmented dataset to ensure a fair comparison.

\subsection{Model Selection}

The first model we have chosen is mT5-small \citep{xue-etal-2021-mt5} of Google which is another sequence-to-sequence Transformer model that is trained on an extensive dataset called mC4 that covers 101 languages, including Bangla and has approximately 300M. It follows the T5 architecture \citep{T5}, featuring an encoder–decoder structure of typical Transformer models. This architecture gives mT5 its “Text-to-Text Transfer Transformer” designation, meaning it converts any NLP task into a text-to-text format — taking text as input and generating text as output. This allows for a multitude of tasks to be performed by the same model with the same hyperparameters - sentence-to-gloss translation being one that mT5 excels at. Furthermore, mT5 is already pretrained on the Bengali language and is thus knowledgeable on Bengali sentence structures and nuances. So this knowledge can be leveraged to better aid a low-resource task like sign language sentence to gloss conversion. Additionally the smaller parameter size enables training even with our limited GPU resources. Taking into account all these factors, along with its solid performance when finetuned for downstream tasks, mT5 inherently feels like a great option for our specific task.

The next selected model is mBART-50 which is an extension of BART (Bidirectional and Auto-Regressive Transformer). It is a multilingual denoising sequence-to-sequence model trained on diverse monolingual corpora across many languages, including low-resource ones such as Bangla, and has a parameter count of approximately 610M \citep{liu-etal-2020-multilingual-denoising}. Due to its multilingual pretraining, mBART-50 effectively transfers knowledge from high-resource to low-resource languages, which results in improved translation quality whether it be supervised or unsupervised. As such, it is particularly suitable for Bangla text-to-gloss translation.


Another model that was useful for our task was NLLB-200-1.3B. \citet{costa2022no} ventured heavily upon this robust multilingual translation model which is capable of handling over 200 languages and thousands of language pairs, including many low-resource ones. It uses a Mixture-of-Experts (MoE) architecture for adaptable multilingual translation. Though its large size risks overfitting on limited data, Parameter-Efficient Fine-Tuning (PEFT) mitigates this, enabling NLLB-200 to generalize well even with small low-resource datasets—ideal for our Bangla text-to-gloss task.


\begin{table*}[htbp]
  \centering
  \small
  \begin{tabular}{@{} l l c c c c c @{}}
    \toprule
      \textbf{Model} & \textbf{Dataset} & \textbf{BLEU-1} & \textbf{BLEU-2}
        & \textbf{BLEU-3} & \textbf{BLEU-4} & \textbf{COMET} \\
    \midrule
      mT5-small      & Base Dataset (1K)      & 51.79 & 36.26 & 25.73 & 17.56 & 0.8564 \\
      mT5-small      & Augmented Dataset (4K) & 55.33 & 40.11 & 29.06 & 20.28 & 0.8602 \\
      mBART-50       & Base Dataset (1K)      & 56.65 & 40.96 & 29.36 & 21.06 & 0.7829 \\
      mBART-50       & Augmented Dataset (4K) & 61.09 & 46.01 & 35.09 & 27.31 & 0.8261 \\
      GPT-4.1-nano   & Augmented Dataset (4K) & 57.11 & 41.50 & 31.11 & 22.42 & 0.8913 \\
      NLLB-200-1.3B           & Base Dataset (1K)      & 67.99 & 40.82 & 26.49 & 16.67 & 0.907 \\
    \bottomrule
  \end{tabular}
  \caption{Model performance comparison using BLEU and COMET metrics on base (1K) and augmented (4K) datasets.}
  \label{tab:metrics}
\end{table*}

\subsection{Model Fine-tuning}

\subsubsection{Fine-tuning mT5 for Bangla Gloss Generation}

According to \citet{xue-etal-2021-mt5} the mT5-small model was pre-trained on a new Common Crawl-based dataset covering 101 languages and has 300M parameters. Therefore the model already understands the underlying structure behind Bangla words and sentences, allowing us to specialize it in our Bangla sentence to gloss generation tasks. We load our Bangla-sentence and gloss pairs with a batch size of 16  because higher batch sizes like 32 and 64 would require a lot of memory for computation. We set our learning rate to 0.001 with warm-up steps set to 50 allowing the model to gradually linearly increase its learning rate to 0.001, preventing large initial updates to the weights and biases which can destabilize the model. Epoch was selected at 20 with early stopping implemented, to prevent overfitting and ensure optimal model selection. Additionally, we also used AdamW optimizer for better regularization during training.

\subsubsection{mBART-50 Fine-tuning Parameters}

The mBART-50 model was pretrained on a diverse set of 50 languages which included Bengali and was primarily developed for  fine tuning on translations tasks \citep{mbart50}. Based on this the model was fine tuned on the following hyper-parameters. An effective batch size of 16 was selected  with the learning rate set to max 3e-5 with warm-up steps set to 300. An epoch of 20 was selected with early stopping implemented.

\subsubsection{Parameter-efficient Fine-tuning of NLLB-200}

NLLB-200 was primarily optimized for machine translation tasks. There are three variants of this model: NLLB-200-distilled-600M, NLLB-200-distilled-1.3B and NLLB-200-3.3B. The models are pre trained on a massive multilingual corpus making them eligible for the Bangla sentence to gloss translation task. We evaluated the performance of the NLLB-200-distilled-1.3B model to assess how a billion-parameter architecture would perform on our dataset. However, conducting a full parameter finetuning of a 1.3B model would require significant computational resources. Therefore, we applied Parameter-Efficient Fine-Tuning (PEFT) techniques, specifically Low-Rank Adaptation (LoRA) which enables effective model adaptation while maintaining computational efficiency \citep{LoRa}. LoRa provides extra trainable parameters which are known as low rank matrices, while preserving the original pre-trained weights thus the original pre-trained knowledge is less modified. Due to our dataset being relatively small we set the rank parameter (r) to 8, which determines the dimensionality of the low-rank matrices and the LoRA. The alpha parameter was set to 16 for a scaling factor 2 so that the learned weight updates are doubled before being added to the original weights, thus allowing the pre-trained model to learn task specific patterns. A dropout rate of 0.05 was applied specifically to LoRA layers to provide regularization and prevent overfitting of the adapter weights. The target module section included q\_proj, k\_proj, v\_proj, out\_proj of the self-attention layers, and fc1, fc2 in MLPs. This allows the model to adapt in terms of both context understanding and feature transformation for gloss sequence generation. The batch size was set to 1 with gradient accumulation steps of 24, creating an effective batch size of 24. Gradient accumulation maintains training stability and enables effective parameter updates despite the small per-device batch size. These steps were mainly taken due to memory limitations, as larger batch sizes would exceed available VRAM. The learning rate was set to 3e-4 with the training epochs set to 5 and mixed precision training (FP16) was enabled to reduce memory usage and accelerate training while maintaining numerical stability. Despite these efforts, we were only able to run the model on the manual annotated part of our dataset.

\subsection{Model Fine tuning Results}

Table~\ref{tab:metrics} shows the effectiveness of our data augmentation strategies for Bangla text to gloss translation. Among these models, mBART-50 achieved the best performance on both datasets, with the augmented dataset (4K samples) showing consistent improvements across all BLEU metrics compared to the professional-only dataset of 1K samples. The model achieved a BLEU-4 score of 27.31 and COMET score of 0.8261 on the augmented dataset. mT5 also benefited from data augmentation, with BLEU-4 improving from 17.56 to 20.28 and achieving a COMET score of 0.8602 on the augmented dataset. Unfortunately, NLLB-200 could not be evaluated on the augmented dataset due to computational constraints, though it showed promising results on the manual annotated dataset of 1000 Bangla sentences with a COMET score of 0.907. We also finetuned GPT-4.1-nano to compare its performance to the other models, and it delivered strong results on the augmented dataset, achieving a BLEU-4 score of 22.42 and the COMET score of 0.8913  The consistent performance gains across models validate our multi-faceted augmentation approach combining RAG-based retrieval, rule-based tense conversion, and masking techniques.


\subsection{Comparative Results with RWTH-PHOENIX-2014T}

For contextual comparison, we also evaluate on the RWTH-PHOENIX-2014T dataset, a German Sign Language benchmark widely used in sentence-to-gloss translation. Although direct comparison is not meaningful due to language and domain differences, similar evaluation settings allow us to observe that our models achieve comparable translation consistency within the Bangla domain. From Table~\ref{tab:bleu_comet_scores}, we can observe that the mBART-50 model achieved high BLEU-4 and comet score of 27.31 and 0.8261 on our dataset, compared to RWTH-PHOENIX-2014T's 21.49 and 0.6673 for mBART-50. Similarly, for mT5, our dataset has also shown BLEU-4 and COMET scores of 20.28 and 0.8602 compared to RWTH-PHOENIX-2014T's 18.73 and 0.6205. 

\begin{table}[htbp]
  \centering
  \small
  \setlength{\tabcolsep}{3.5pt} 
  \resizebox{0.95\columnwidth}{!}{
    \begin{tabular}{@{} l l c c c c c @{}}
      \toprule
        \textbf{Dataset} & \textbf{Model} & \textbf{BLEU-1} & \textbf{BLEU-2} 
          & \textbf{BLEU-3} & \textbf{BLEU-4} & \textbf{COMET} \\
      \midrule
        \multirow{2}{*}{\makecell{RWTH\\PHOENIX}}
          & mBART-50 & 48.49 & 35.01 & 26.85 & 21.49 & 0.6673 \\
          & mT5-small & 54.79 & 36.47 & 25.42 & 18.73 & 0.6205 \\
        \midrule
        \multirow{2}{*}{\makecell{Augmented\\Dataset (4K)}}
          & mBART-50 & 61.09 & 46.01 & 35.09 & 27.31 & 0.8261 \\
          & mT5-small & 55.33 & 40.11 & 29.06 & 20.28 & 0.8602 \\
      \bottomrule
    \end{tabular}
  }
  \caption{Evaluation results on RWTH-PHOENIX-2014T and our datasets.}
  \label{tab:bleu_comet_scores}
\end{table}

\section{Conclusion}
In this paper, we have introduced Bangla-SGP, a comprehensive, high-quality Bangla Sentence Gloss pair dataset, which will contribute greatly to this low-resource domain while also laying the groundwork for promising future research in continuous sign language recognition and translation. Furthermore, through our proposed novel RAG-based pipeline, we have introduced and proved the effectiveness of utilizing RAG in generating high-quality synthetic gloss sequences. Recognizing the potential of data augmentation in this low-resource medium where expert manual annotation is quite costly, we explored and devised various other data augmentation strategies along with our proposed RAG-based scheme. Our findings demonstrate that data augmentation is a very effective and viable solution for expanding gloss resources as well as overcoming the limitations of low-resource settings.

\section{Dataset Availability Statement}
The Bangla-SGP dataset will be made publicly available upon publication through our official repository at GitHub. The dataset will be released under the CC BY-4.0 license and all augmentation rules, and documentation will be included in the repository.

\section{Discussion}
\subsection{Research Limitations}
While our work lays down the foundation for future work on Continuous Bangla Sign Language Recognition and Translation, it also comes with several limitations that need to be addressed. Firstly, although our dataset of 4000 samples is a valuable extension to such a low-resource setting, such as sign language NLP tasks, it is still relatively limited to ideally train modern transformer architectures to perfectly capture the nuances and morphological patterns that come with converting a Bangla sign language sentence to its corresponding gloss representation. Secondly, manual annotation of Bangla sentence-gloss pairs by an expert is an expensive procedure due to the limited availability of Bangla sign language experts. Due to this, we were only able to cooperate with one expert in our work, so our dataset may include signer bias. Thirdly, sign language is a complex form of communication that is not restricted to hand movements; rather, it also involves facial expressions and body movements to convey different words, emotions, and tones. Glosses, however, as an intermediary, cannot capture facial expressions, and so our dataset lacks a video component to capture these aspects. Additionally, Bangla sign language is limited in vocabulary, due to which many words that the hearing community uses may not have a sign language/gloss representation. So, signers usually break down that word into a set of glosses to express the word. For example, the word (shoptaho) ‘week’ is broken down into (shat din) ‘seven days’ by signers. As such, our dataset needs to be expanded upon with more such examples for machine translation to capture these special relationships. Finally, names and landmarks do not typically have a dedicated single sign; they are usually spelled out using alphabet-level signing. Glosses like these don't have a specific representation in our dataset which may cause challenges during the training of more advanced sign language translation systems. Finally, due to constraints in computational resources, we were not able to evaluate NLLB-200-1.3B on our augmented dataset. As a result, we cannot empirically verify whether exposure to the augmented sentence–gloss pairs would yield the same improvements in the evaluation scores.

\subsection{Future Works}

The presented dataset is the Phase 1 of a multimodal BdSL dataset, consisting only of the gloss sequences of spoken Bangla text. We acknowledge that non-manual markers such as facial expressions, head movements and torso movements play a critical role in sign language grammar and meaning. In Phase 2, we will collect synchronized sign video samples of the Bangla sentences and add annotations for non-manual markers with the help of multiple signers. Subsequently, through inter-annotator agreement checks, we aim to develop a comprehensive dataset that supports research on both lexical and multimodal aspects of sign language. 

We plan to use the extensive dataset for creating a pipeline that generates 3D sign language representations from Bangla text. The planned pipeline will build upon the gloss generation process from Bangla text and involve constructing a gloss video dictionary that consists of isolated gloss, which serve as the key, and their corresponding RGB video representations as their value. From these videos, through the use of State-of-the-art 3D human reconstruction architectures such as SMPLer-X, generation of 3D human meshes as obj files is possible on a frame-by-frame basis \cite{cai2024smplerxscalingexpressivehuman, smplx}. These meshes are mapped back to their respective entries in the gloss video dictionary, and the resulting series of obj files can be used to produce a continuous 3D animation of the signed sentence. We tried exploring this pipeline on a preliminary level but several challenges remain, including the lack of a comprehensive sign video dataset and unstable transitions between consecutive signs during animation. We were unable to solve these issues, which we will also leave as future work or for other researchers to address.

\subsection{Ethical Consideration}
All of the gloss sequences of the Bangla sentences was annotated with explicit consent from a certified Bangla Sign Language (BdSL) interpreter. We have obtained informed consent from them and they are aware of their data being used for academic research purposes. 
\section{Bibliographical References}\label{sec:reference}

\bibliographystyle{LREC2026/lrec2026-natbib}
\bibliography{LREC2026/lrec2026-example}

\label{lr:ref}
\bibliographystylelanguageresource{LREC2026/lrec2026-natbib}

\end{document}